\documentclass[sigchi, fontsize=10pt]{acmart}
\renewcommand\footnotetextcopyrightpermission[1]{} 
\usepackage{tikz}
\AtBeginDocument{%
  \providecommand\BibTeX{{%
    \normalfont B\kern-0.5em{\scshape i\kern-0.25em b}\kern-0.8em\TeX}}}

\setcopyright{none}

\begin{document}

\title{Multi Document Reading Comprehension}

\author{Avi Chawla}
\email{avi.chawla.cse16@iitbhu.ac.in}
\affiliation{%
  \institution{Indian Institute Of Technology (BHU) Varanasi }
  \city{Varanasi}
  \state{India}
}









\begin{abstract}
Reading Comprehension (RC) is a task of answering a question from a given passage or a set of passages. In the case of multiple passages, the task is to find the best possible answer to the question. Recent trials and experiments in the field of Natural Language Processing (NLP) have proved that machines can be provided with the ability to not only process the text in the passage and understand its meaning to answer the question from the passage, but also can surpass the Human Performance on many datasets such as Standford's Question Answering Dataset (SQuAD) \cite{DBLP:journals/corr/RajpurkarZLL16}. This paper presents a study on Reading Comprehension and its evolution in Natural Language Processing over the past few decades. We shall also study how the task of Single Document Reading Comprehension acts as a building block for our Multi-Document Reading Comprehension System. In the latter half of the paper, we'll be studying about a recently proposed model for Multi-Document Reading Comprehension — RE\textsuperscript{3}QA \cite{DBLP:journals/corr/abs-1906-04618} that is comprised of a Reader, Retriever, and a Re-ranker based network to fetch the best possible answer from a given set of passages.

\end{abstract}



\maketitle
\pagestyle{plain}

\section{Introduction}
The ability to read and understand the unstructured text, and then answer the questions about it is an ordinary skill among literate humans. But for machines, it is not easy.  
Teaching our machines to read natural human language has always been a long-standing goal expected from Natural Language Processing \cite{10.1111/j.1460-2466.1960.tb00541.x}. Throughout many years of efforts, researchers in the field of NLP have used more and more powerful NLP tools to analyze and understand different aspects of human texts. A point that arises here is why we should build these tools, and what is the need behind building such tools? Basically, how can a machine's reading ability be evaluated is the primary question that comes to mind. The answer to the question lies in the task of Reading Comprehension. Just like human knowledge is tested based on asking questions, we should similarly ask machines about what it has understood from the text.\\

\begin{center}

    \includegraphics[width=9cm]{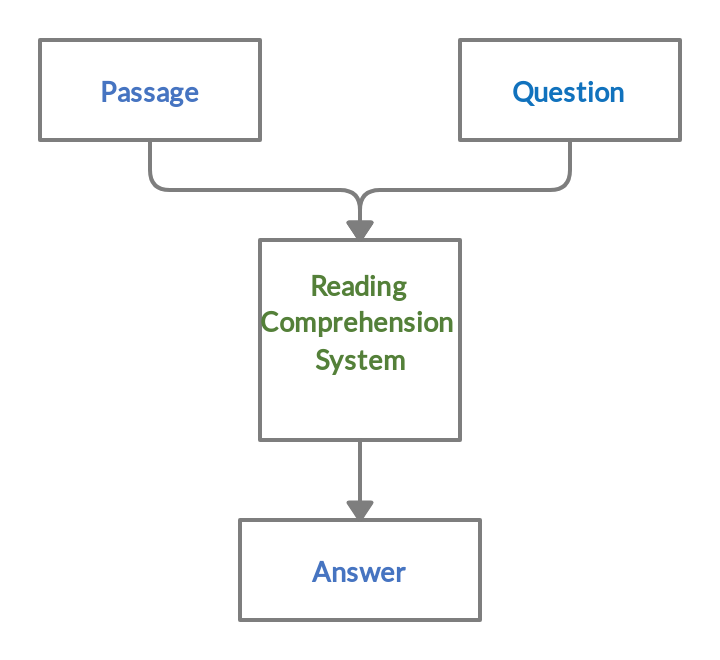}
    \captionof{figure}{Diagram representing a basic Reading Comprehension System}
    
\end{center}

Reading Comprehension is not a new topic. As earlier as 1977, Researchers already realized the importance of Reading Comprehension. \emph{Wendy Lehnert} \cite{nn-1980-process} stated the following line in his book, "The Process of Question Answering":\\

\emph{\Large{"Only when we can ask a program to answer a question about what it reads will we be able to begin to access that program's comprehension."}}\\

Reading Comprehension can be cast as a Question Answering task, which is also an application of Natural Language Processing(NLP) wherein a model is supposed to answer a question put up by the user. The machine answers the question by arbitrarily using some complex, unstructured, semi-structured knowledge bases and returns the correct answer to its user. Question Answering is another of the big successes in using Deep Learning inside Natural Language Processing, and it's a technology that has some obvious commercial usage. \\

Question Answering systems can be classified into two types:
\begin{enumerate}
    \item \emph{Closed-Domain Question Answering} \cite{articlecdm}: This is an easier type of Question Answering wherein the question belongs to a particular domain/topic. Questions in this domain are generally asked for descriptive or factoid information.
    \item \emph{Open-Domain Question Answering } \citep{DBLP:journals/corr/ChenFWB17} : In contrast to the category mentioned above, here the question can belong to any class and thus, can pose anything in the world. These systems generally require much data to answer the question accurately, e.g., Wikipedia Dump.
\end{enumerate}

Here, the goal is to understand the passage and answer the question from the passage. NLP systems in such a scenario are expected to leverage the power of Natural Language Understanding, Natural Language Inference or Entailment, and Information Retrieval, etc. Thus, reading comprehension is challenging for us. \\

There are two main paradigms for Question Answering. These two paradigms cover the majority of Question Answering Systems that we use or build today. These are:
\begin{enumerate}
    \item \emph{IR-based Approaches:} These systems concentrate on finding an answer to a question by looking at strings of text. Google Search is a common example of this approach. \cite{Abbasiyantaeb2020TextbasedQA} and \cite{Gupta2012ASO} also presents a detailed survey on the same.
    \item \emph{Knowledge-based or Hybrid Approaches:} These systems build the answer from understanding a part of the text. Natural Language Understanding plays a crucial role in building such systems. Examples include IBM Watson, Apple Siri. A detailed study on such systems is given by \cite{DBLP:journals/corr/abs-1903-02419} and \cite{Zhu2020KnowledgebasedQA}
\end{enumerate}

A comprehensive study between the IR-based approaches and Knowledge-based approaches is presented in \cite{articlestudy}.
Question asked to a Question Answering can be of many times, ranging from Factoid Questions to Complex Narrative Questions. A few types of Question Categories are shown in Table 1.

\begin{table}[t]
\resizebox{0.48\textwidth}{!}
{%
\begin{tabular}{@{}r|r|r@{}}
\toprule
\textbf{S.No.} & \textbf{Question Category} & \textbf{Description} \\
\midrule
1 & Verification Questions & Intends a 'Yes' or 'No' answer. \\
2 & Factoid-type Questions & What, Which, When, Who or How? \\
3 & Casual Questions & Why or How? \\
4 & Hypothetical Questions & What would happen if ...? \\
5 & Complex Questions & What are the reasons for ...? \\
6 & Definition Questions & What is the definition of ...? \\
7 & Quantification Questions & How much is ...? \\

\bottomrule
\end{tabular}%
}
\caption{A few question types generally posed by humans.}\label{tab:all-results}
\end{table}

\section{History of Question Answering}
Two of the oldest Question Answering system were called BASEBALL \citep{10.1145/1460690.1460714} and LUNAR \citep{articlelunar}. \\

BASEBALL was designed way back in 1961 to answer questions related to US Baseball League for little more than a year. It accepted question in English language and answered from an already stored knowledge base.\\

On the other hand, LUNAR was developed in 1971 and was capable of answering questions about moon rocks and soil that the Apollo 11 Mission had gathered in July 1969. \\

Both the systems, BASEBALL and LUNAR, were Closed-Domain Question Answering systems as they answered questions related to a particular topic only and not about anything in the world at that time. \\

SHRDLU \cite{Ontan2018SHRDLUAG} was another such domain-restricted question answering system that was developed in late 1968. Researchers also called this system the "Blocks World" wherein a user interacted in English with SHRDLU to move various objects in the "Blocks World." It comprised of different basic geometrical objects such as Cone, Cylinder, Cube, etc. \\

\begin{center}

\includegraphics[ width=6cm]{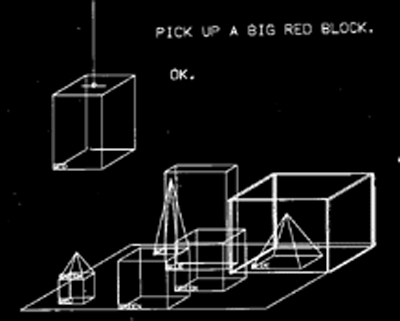}
\captionof{figure}{SHRDLU system working}

\end{center}

Simmons \cite{10.1145/363707.363732} also did an exploration of answering questions using dependency parses matching techniques of question and the answer. This is demonstrated in the figure below.

\begin{center}

\includegraphics[ width=6cm]{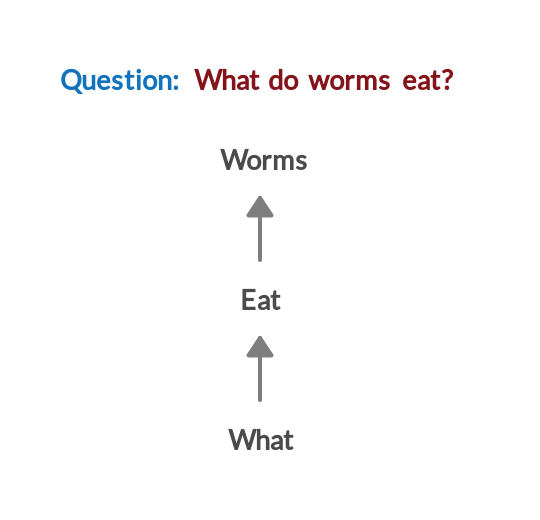}
\captionof{figure}{Question}

\includegraphics[ width=6cm]{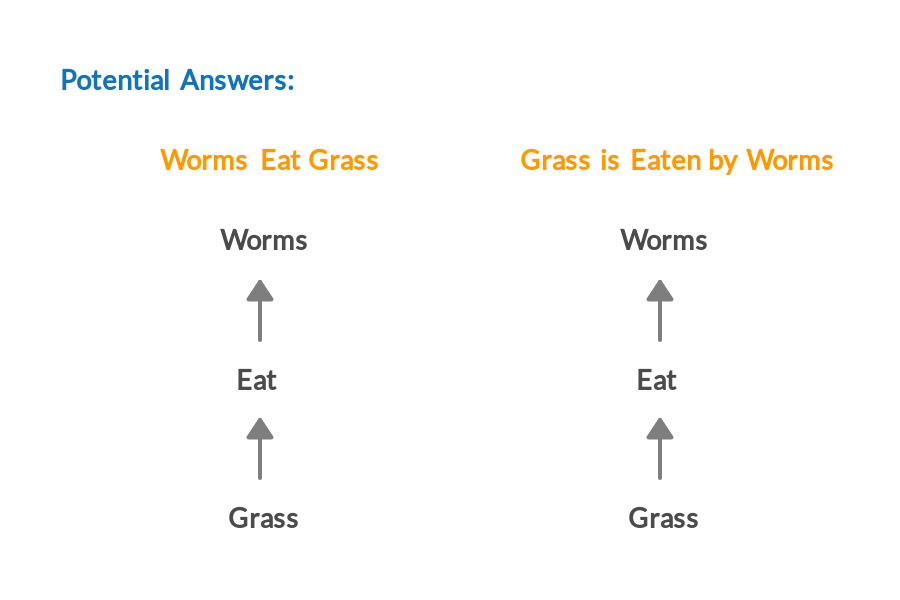}
\captionof{figure}{Potential Answers}

\end{center}

Considering the above example, the system took the question and parsed it into a dependency format to establish a relationship between the words. The next step was to find a matching dependency structure in the extensive database of answers. Figure 4 above shows some of the potential answers that the system fetched through it's matching algorithm.\\

At the start of this decade, IBM's Watson \cite{articleibm} \cite{6622216} came at the forefront of a new era of Cognitive Computing, beating humans in the very famous Jeopardy Challenge, back in 2011. Cognitive Computing is a very different type of computing that is very different from the programmed systems that have preceded it. With today's world of big data and the need for more complex evidence-based decisions, approaches such as Decision Tress \cite{inbook} or Tabulated Answer Retrieval based approach often breaks or fails to keep up with the available information. Cognitive Computing enables machines to create a new kind of value and find answers through insights drawn from vast amounts of data. Watson and its Cognitive capabilities adhere to some of the critical abilities and elements of human expertise, leading to a system that solves a problem as a human does. When we seek to understand something and try to make a decision, we go through four key steps:
\begin{enumerate}
    \item We \emph{observe} the visible phenomena and bodies of evidence.
    \item We mix what we know and see to \emph{interpret} its meaning and generate a hypothesis about what it means.
    \item We \emph{evaluate} which hypothesis is right and wrong.
    \item Finally, we \emph{decide}, choosing the option that seems best and acting accordingly.
\end{enumerate}
Just as humans become experts by going through Observation, Evaluation, and Decision-making, Cognitive Systems like Watson use similar processes to explain the information they read. Watson can do all this at a massive speed and scale.

\section{Evolution}

Although the importance of Reading Comprehension has been recognized for over 45 years, Reading Comprehension is a very new domain to explore in the Natural Language Processing Research Community. This is evident from the following information:

\subsection{Pre-2015}
In this period, i.e., before 2015, we didn't have any statistical NLP systems capable of reading a simple passage and answering questions from it. Considering the Dataset aspect of this period, before 2015, we hardly had two datasets consisting of passage, question, and answer pairs. There were:
\begin{enumerate}
    \item MCTest \cite{richardson-etal-2013-mctest}: 2600 pairs of question, passage and answer.
    \item ProcessBank \cite{berant-etal-2014-modeling}: 500 pairs of question, passage and answer.
\end{enumerate}
If we look at the system or model aspect of this period, most of the models built for experimentation in this period were hand-built systems like linear classifiers with some commonly used linguistic features known at that time. 

\subsection{Post-2015} 

In this period, some significant changes came in the field of Reading Comprehension. Talking about the Dataset aspect of this period, post-2015, we had enormous datasets with us to build Reading Comprehension or Question Answering Systems. These datasets were collected from various sources like News Articles or Wikipedia or children's books. A few examples of these datasets are:

\begin{enumerate}
    \item CNN Daily Mail \cite{DBLP:journals/corr/HermannKGEKSB15}
    \item SQuAD \cite{DBLP:journals/corr/RajpurkarZLL16}
    \item LAMBADA \cite{DBLP:journals/corr/PapernoKLPBPBBF16}
    \item Who Did What (WDW) \cite{DBLP:journals/corr/OnishiWBGM16}
    \item Children's Book Test (CBT) \cite{Hill2016TheGP}
    \item MSMACRO \cite{DBLP:journals/corr/NguyenRSGTMD16}
    \item Maluuba News QA \cite{DBLP:journals/corr/TrischlerWYHSBS16}
    \item TriviaQA \cite{DBLP:journals/corr/JoshiCWZ17}
    \item WikiReading \cite{hewlett-etal-2016-wikireading}
    \item SearchQA \cite{DBLP:journals/corr/DunnSHGCC17}
    \item CMU Race \cite{lai-etal-2017-race}
\end{enumerate}
One common thing among all these datasets was that they all contained more than 1 lakh pairs of Passage, Question, and Answer. If one looks at the system side of this period, a revolution in the model designs and architectures took place. We saw that all the hand-built systems of the Pre-2015 period evolved into robust end-to-end neural network architectures. Not only did it outperform the earlier-used linear classifier based models but also started achieving near-human performance on many datasets mentioned above.

\section{Applications of Question Answering in Daily-Life}

These days it seems like every major tech company either has or is working on a Virtual Assistant to enhance the customer experience. Researchers in Artificial Intelligence have been dreaming of computers we could talk to for decades now. It appears that now, we have started to get to the point where these friendly, helpful voices are beginning to feel unavoidable. Recent innovations in the field of Artificial Intelligence have led to the development of numerous Virtual Assistants that help us significantly in our day to day lives. These Virtual Assistants are based on Machine Learning Techniques like Question Answering Systems. Each Virtual Assistant and its closely related chat-bot systems run just a bit differently, but they all have the same result, making our lives much more comfortable. These chat-bots eliminate the availability of humans on a variety of tasks like the management of a company's Customer Assistance portal, guiding school students with their study plan, and many more. Virtual Assistants and the technology around them are helping us do the things we do every day just quicker and more comfortable. \\
In recent times, these technologies have now even surpassed the Question Answering System. These systems have become proficient in making real-time conversations with the users and asking follow-up questions to understand better what we want. Some popularly known examples of Virtual Assistants that we use daily are Google Assistant by Google, Siri by Apple, Alexa by Amazon, and Cortana by Microsoft. A comprehensive overview of such Virtual Assistants and their working is presented in \cite{inbookassis}.

\section{Related Works to Question Answering}

Machine Reading Comprehension is a very old problem in the field of NLP. With massive collections of full-text documents, i.e., the Web, simply returning relevant documents is of limited use. So instead, what we look for these days is the direct answer to our questions. In this regard, many recent Natural Language Processing Systems have heavily leveraged the power of Deep Learning and other massive NLP architectures for providing a solution to the Question Answering problem of NLP. 

\subsection{Statistical Question Answering}

A majority of basic and traditional approaches to answering question answering involve techniques such as rule-based algorithms or the employment of linear classifiers on a set of hand-engineered features. Lynette Hirschman proposed Deep Read \cite{hirschman-etal-1999-deep} built on a training corpus of short stories in the books of students of 3\textsuperscript{rd} to 6\textsuperscript{th} grade. The task aimed at reading a story and answering a question based on that. The proposed system made use of a simple bag-of-words approach technique for matching the question with the passage to find the most relevant answer. \cite{richardson-etal-2013-mctest} proposed two baselines for the task of Question Answering. One of them uses used some simple techniques such as a sliding window to match the bags-of-words from the passage. The other made use of word-distances between words present in the question and in the document. Some statistical approaches have also focused on learning a structured representation of the entities in the data and storing the relations in the document as a knowledge-base \cite{berant-liang-2014-semantic}. The next step focuses on the conversion of the question to a structured query. This allows the model to match the content of the question to that of the knowledge-base. Other statistical approaches have revolved around the usage of semantic features and the syntactic features of words such as POS \cite{wang-nyberg-2015-long}. One of the finest and competitive baseline came through the efforts of \cite{chen-etal-2016-thorough}. They proposed an approach using a set of carefully crafted lexical, syntactic, and word order features of the Question and Answering data. Christopher Burges \cite{richardson-etal-2013-mctest} proposed a challenge problem for Artificial Intelligence and released a Reading Comprehension training corpus called the MCTest \cite{richardson-etal-2013-mctest}.

\begin{table*}[t]
{
\centering
\resizebox{\textwidth}{!}
{%
\begin{tabular}{@{}r|r|p{0.35\textwidth}|r|r|r@{}}
\toprule
\textbf{S.No.} & \textbf{Dataset Name} & \centering\bfseries \textbf{Description} & \textbf{Release Year} & \textbf{Training Samples} & \textbf{Testing Samples} \\
\midrule
1 & CliCR \cite{DBLP:journals/corr/abs-1803-09720} & Created from reports of Clinical Cases & 2018 & 91,344 & 7,184 \\
2 & CNN \cite{DBLP:journals/corr/HermannKGEKSB15} & Created from News Articles & 2016 & 380,298 & 3,198 \\
3 & Daily Mail \cite{DBLP:journals/corr/HermannKGEKSB15} & Created from News Articles & 2016 & 879,450 & 53,182 \\
4 & CoQA \cite{reddy-etal-2019-coqa}  & Meant for Building Conversational Question Answering system & 2019 & 1,27,000* & - \\
5 & MS MARCO \cite{DBLP:journals/corr/NguyenRSGTMD16} & Obtained from Bing's Search Results & 2016 & 1,010,916 & - \\
6 & NewsQA \cite{DBLP:journals/corr/TrischlerWYHSBS16} & Created from CNN News Articles & 2016 & 119,633 & - \\
7 & RACE \cite{lai-etal-2017-race} & Created from English examinations in China & 2016 & 87,866 & 4,934 \\
8 & SQuAD \cite{DBLP:journals/corr/RajpurkarZLL16} & Created from Wikipedia Data & 2016 & 129,941 & 5,915 \\

\bottomrule
\end{tabular}%
}
\caption{A few question types generally posed by humans.}\label{tab:all-results}
}
\end{table*}

\subsection{Neural Question Answering} 

Deep Learning has always been the key to many extraordinary advancements done in the field of NLP. Neural Models find a vast application in the field of Question Answering also. For instance, the use of Neural-attention models has been widely applied in recent times for machine comprehension in NLP. \cite{DBLP:journals/corr/HermannKGEKSB15} proposed an Attentive Reader model in addition to a large cloze-style question answering dataset CNN/Daily Mail that led the foundation of many Deep Learning-based architectures in the future. Another dataset was released that originated from the Children's book data \cite{Hill2016TheGP}. As a baseline, they proposed a window-based memory network. The use of pointer networks has also gained popularity for answer prediction \cite{DBLP:journals/corr/KadlecSBK16}. \cite{DBLP:journals/corr/SordoniBB16} also made attempts in using the neural attention model for the task of machine comprehension. One major revolution in Neural QA came about through the release of the SQuAD dataset \cite{DBLP:journals/corr/RajpurkarZLL16}, which comprises of Question, Passage and Answer triplets. The answer is always a continuous span to words in the passage, and the questions appear to be more realistic in this dataset. Since then, many attempts have been made to build a Question-Answering system on SQuAD.  \cite{DBLP:journals/corr/WangJ16a} proposed an end-to-end neural network-based approach that comprises of a Match-LSTM encoder. \cite{DBLP:journals/corr/YuZHYXZ16} make use of a dynamic chunk reader, which is a neural reading comprehension model aimed at the extraction of a set of potential answers from the passage and then ranks these answers to answer the question precisely. 

The scope of Question Answering is not restricted to only text-based Question Answering. Many attempts have also been made in the past to propose models for Visual Question Answering. For instance, \cite{DBLP:journals/corr/LuYBP16} proposed a hierarchical co-attention model for this task. The performance of this model on the COCO-VQA dataset \cite{DBLP:journals/corr/AntolALMBZP15} achieved the state of the art results. This followed the development of another VQA model that made use of a co-attention mechanism to compute a conditional representation of the image given the question and vice versa \cite{Lu2016HierarchicalQC}.
Inspired by the above idea, \cite{DBLP:journals/corr/AntolALMBZP15} proposed a Dynamic Co-attention Network (DCN) and an year later a Dynamic Co-attention Network+ (DCN+) \cite{DBLP:journals/corr/abs-1711-00106}. The architecture comprises of a Co-attentive encoder and a Dynamic Decoder. The answer is predicted with the help of a Pointer Network \cite{Vinyals2015PointerN}. 

In recent times, NLP has witnessed a considerable work in the direction of generating Contextualised Word Embeddings (CWEs) \cite{devlin2018bert} \cite{Peters:2018} \cite{akbik2018coling}. CWEs have been inspired by the idea of Sense Embeddings, which suggests using different Word Embeddings for each sense of the word. Word Embeddings models such as Word2Vec \cite{41224} and Glove \cite{pennington2014glove} ignore polysemy of words and provide only a single embedding for each word. In contrast, CWEs provide a separate embedding for each word's occurrence in the context it appears. One such revolution in the field of CWEs can be realized by the efforts made by the \cite{devlin2018bert} and their proposed BERT model. The model is based on a Transformer \cite{DBLP:journals/corr/VaswaniSPUJGKP17} architecture that makes use of Attention Mechanism to find the Contextualised Embeddings for words. Recent approaches in Question Answering have also focused on using these models in place of the previously used Static pre-trained Embeddings. 

\section{Datasets}

In the past few years, researchers in the field of NLP have put in many efforts of curate Reading Comprehension datasets. Most of these datasets are composed of a Passage, Question, and Answer triplets and consists of approximately one lakh training examples. A brief overview and statistics of some of the popularly used datasets for Reading Comprehension are given in Table 2.

\begin{figure*}[t]
    \centering
        \fbox{\includegraphics[width=\textwidth]{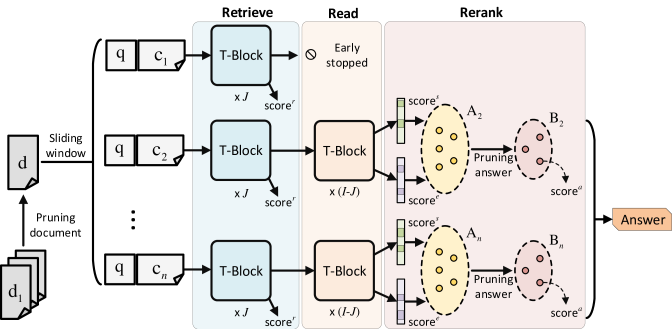}}
    \caption{Figure shows the flow of final answer prediction from the RE\textsuperscript{3}QA approach to Multi-Document Reading Comprehension}
\end{figure*}

\section{Multi Document Reading Comprehension}

The task of natural Reading Comprehension is of minimal practical use. It does not allow the system to search in multiple documents to search for an answer. The approaches discussed in the Related Work Section primarily focus on such a scenario of Question Answering. In contrast, Multi-Document Reading Comprehension refers to the task of finding the correct answer to a question by searching in more than one documents. This task has more applications as an end product than the Single document Reading Comprehension. 

Most of the approaches proposed in this area have heavily relied on the working mechanism of a natural Reading Comprehension system. This system functions very in a fashion remarkably similar to that of the natural one. It just adds an extra step of ranking the answers obtained from various sources. A basic pipeline to this task is explained below in the following steps:

\begin{enumerate}
    \item \emph{Document Selection}: Out of the myriad of documents available in the knowledge base, this step involves the selection of top \emph{n}-documents that match to the question.
    \item \emph{Finding Candidate Answers}: In this step, the model tries to find some potential answers to the question from the selected sets of documents.
    \item \emph{Answer Ranking and Selection}: This step aims to rank the answers extracted from the previous step according to their relevance to the question and return the best answer to the question.
\end{enumerate}

In recent times, researchers working in the area on Multiple Document Reading Comprehension have gained considerable success in building efficient systems that can find answers to question hidden in volumes of data \cite{DBLP:journals/corr/abs-1811-11374} \cite{DBLP:journals/corr/abs-1906-04618}. These systems leverage the power of highly efficient and effective algorithms and architectures at each step. \cite{choi-etal-2017-coarse} focused on the construction of a coarse-to-fine grained framework that answers the question from a retrieved document summary. \cite{inproceedingslin} proposed a pipeline system consisting of a paragraph selector and a paragraph reader. \cite{DBLP:journals/corr/abs-1805-02220} extended \cite{choi-etal-2017-coarse} and introduced a network for cross-passage answer verification. This paper studies one such recent approach that makes use of the end-to-end network for efficient answer retrieval from multiple documents. This approach is based on Retriever, Reader, and Re-ranker model that works in the exact similar to the pipeline described above. The approach is described in the following section.

\section{RE\textsuperscript{3}QA: Retriever, Reader, and Re-ranker based Question Answering}

Figure 1 gives an overview of the RE\textsuperscript{3}QA model for Multi-Document Reading Comprehension taken from \cite{DBLP:journals/corr/abs-1906-04618}. The model comprises of three sub-networks. These are:
\begin{enumerate}
    \item \emph{Retriever}: Retriever aims to select a few relevant documents from volumes of data available to us as a knowledge base.
    \item \emph{Reader}: Reader's works on the output of Retriever, and it is used to extract candidate answers from the retrieved documents.
    \item \emph{Re-ranker}: Re-ranker aims to re-score multiple candidate answers generated by the Reader and return the best answer to the question.
\end{enumerate}

In contrast to previously adopted approaches, the RE\textsuperscript{3}QA model integrates the Retriever, Reader, and Re-ranker components into a unified network instead of a pipeline and is trained using an end-to-end strategy. One more improvement is brought to the model by providing it Contextualised Word Embeddings (CWEs) of BERT rather than pre-trained Embeddings such as Glove Vectors. The use of BERT has already been found to be very useful on many NLP task, and the use of BERT Contextualised Embeddings in RE\textsuperscript{3}QA is what accounts for its exceptional performance gains. This helps the model better understand the text and establish relationships between the question and the knowledge base to be searched. 

The three sub-networks that govern the entire answer generation process run through a five-staged process. These are described in the following sub-sections:

\subsection{Document Pruning}

This is the first of five steps involved in the answer retrieval process for this model. As the name suggests, this step involves the selection of potential documents from the entire database which may contain the answer to the question. \\

\begin{figure}[h]
    \resizebox{0.48\textwidth}{!}
    {
        \fbox{\includegraphics[width=\textwidth, height = 5cm]{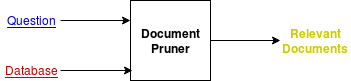}}
    }
    \caption{}
\end{figure}

The input to this step is the Question Q and the set of Documents $D \in \{ D\textsubscript{1}, D\textsubscript{2}, D\textsubscript{3}, \dots D\textsubscript{N} \}.$ Paragraphs in each Document are ranked according to TF-IDF scores with the question and top-K paragraphs are selected for further steps. These paragraphs are then concatenated to form a a single document \emph{d}.

\subsection{Segment Encoding}

As the name suggests, this part involves splitting of the Document d obtained in the previous step into small components. To achieve this, the model slides a window of length \emph{l} with a stride \emph{r} over the Document \emph{d}. This results in the formation of a set of text-segments $C \in \{ C\textsubscript{1}, C\textsubscript{2}, C\textsubscript{1}, \dots C\textsubscript{N} \}.$

In the next step, these segments are individually concatenated with the Question Q and encoded with the help of the BERT model. Input to the Transformer Block is QC = ['[CLS]'; Q ; '[SEP]' ; C\textsubscript{i} ; '[SEP]'] where '[CLS]' is called the Classification Token and '[SEP]' is a token used for separating sentences. The length of vector QC is L\textsubscript{X}. This is then passed into the TransformerBlock which returns I hidden layers. The value of \emph{I} for bert-small is 12 and for bert-large is 24. Each hidden layer is of size L\textsubscript{X}*D\textsubscript{H} where D\textsubscript{H} is the size of hidden layer. 

\begin{center}
    $hi = TransformerBlock(h(i-1)) where i \in [1, I]$ \\
    $hi \in \mathbb{R}\textsuperscript{L\textsubscript{X}*D\textsubscript{H}}$
\end{center}
    
\subsection{Early Stopped Retriever}

The above encoding is very effective and appealing but there is a lot of computational efficiency involved in it as well. This becomes evident from the fact that average number of text-segments C for each question is roughly 20. Each segment is concatenated with the Question Q and passed through atleast 12 hidden layers for encoding it. Each hidden layer is atleast of size 768. The Early Stopped Retriever applies another level of pruning. In a nutshell, it ranks all the text-segments using the hidden representations obtained in Segment Encoding step. This is done using a series of Multi-Layer Perceptron and a self-aligning layer with the hidden states. To train this component, it is trained on an objective function which computes the loss over those ranked segments that don't contain any word of the final answer. From this step, we select \emph{M} text-segments according to the ranks obtained for further processing.

\subsection{Distantly-Supervised Reader}
The aim of the Reader is to generate multiple candidate answers. This is achieved by projecting the last hidden layer of each text-segment obtained in the previous step into two scores. These scores determine the potential start and end of the answer. The objective function defines a loss that is totally similar to that of the Early Stopped Retriever model. The module returns the start and the end position of the preliminary answers from each of the \emph{M} text-segments. These are: A = [A\textsubscript{1}, A\textsubscript{2}, A\textsubscript{3}, \dots, A\textsubscript{M}].

\subsection{Answer Re-ranker}

This module ranks the potential answers from the previous module with the help of their span representations. The objective function defined here is also similar to that described in the earlier modules. This objective function takes into account whether a particular span representation contains at least one token from the original answer or not. Those representations that don't contain any such token are used in the loss function. \\

In the above sub-sections, the objective of each of the three-modules, i.e. the Retriever, Reader and the Re-ranker are defined. Rather than separately training each component, the training strategy involves an end-to-end strategy for training them together. This is done by adding the three loss functions and train the entire network in based on this loss function using Gradient Descent.

\subsection{Datasets}

Experimentation is done on 3 prominent and widely used Question Answering datasets. These are:
\begin{enumerate}
    \item TriviaQA-Wikipedia: This dataset consists of Top-10 search documents from Wikipedia. 
    \item SQuAD Document: It is variant of the original SQuAD dataset. Difference is that it maps the question to the entire Wikipedia Page instead of a specific paragraph.
    \item SQuAD Open: This dataset is basically the entire Wikipedia-domain.

\end{enumerate}

\subsection{Results}

The RE\textsuperscript{3}QA model, when trained with the appropriate data and training strategy, outperforms the previously adopted approaches and achieves state-of-the-art results on the three-datasets written above. The results for each of these models is shown in the following table:

\begin{table}[h]
\resizebox{0.48\textwidth}{!}
{%
\begin{tabular}{@{}r|r|r@{}}
\toprule
\textbf{Dataset} & \textbf{RE\textsuperscript{3}QA Large} & \textbf{RE\textsuperscript{3}QA Base} \\
\midrule
TriviaQA-Wikipedia & 83.0 & 79.9 \\
SQuAD Document &  87.20 & 84.81 \\
SQuAD Open & 50.2 & 48.4 \\
\bottomrule
\end{tabular}%
}
\caption{F1(\%) scores obtained by using the BERT-base and the BERT-large model on each of the three datasets. There are taken directly from \cite{DBLP:journals/corr/abs-1906-04618}.}\label{tab:all-results}
\end{table}

\section{Conclusion}

This paper presented a study on the evolution of Question Answering as an AI problem and its importance in our day to day lives. We also discussed the history of Question Answering over the past few decades and how the advancements in the field have led to the development of plenty of Chat-bots using Question Answering techniques. In addition to this, we discussed two significant periods of Question Answering, i.e., the Pre-2015 and the Post-2015. While coming to the end of our study, we saw one of the recently proposed approaches to Multi-Document Reading Comprehension that made use of a unified network to answer questions by looking into multiple documents at a time. The model discussed is currently the state-of-the-art in this domain on three challenging datasets of Question Answering.

\bibliographystyle{ACM-Reference-Format}
\bibliography{sample-base}

\end{document}